\documentclass{llncs}

\usepackage[english]{babel}

\usepackage[%
rm={oldstyle=false,proportional=true},%
sf={oldstyle=false,proportional=true},%
tt={oldstyle=false,proportional=true,variable=true},%
qt=false%
]{cfr-lm}
\usepackage{varwidth}
\usepackage{graphicx}

\usepackage{paralist}
\usepackage{comment}
\usepackage{latexsym}

\usepackage{cite}

\usepackage[T1]{fontenc}
\usepackage{cmap}
\usepackage{multirow}
\usepackage[math]{blindtext}

\usepackage{csquotes}

\usepackage{microtype}

\usepackage{url}
\makeatletter
\g@addto@macro{\UrlBreaks}{\UrlOrds}
\makeatother

\usepackage{xcolor}

\usepackage[
   bookmarks=false,
   colorlinks=true,
   allcolors=black,
   pdfstartview=Fit,
   breaklinks=true,
]{hyperref}
\usepackage[all]{hypcap}

\usepackage{pdfcomment}

\usepackage[capitalise,nameinlink]{cleveref}
\crefname{section}{Sect.}{Sect.}
\Crefname{section}{Section}{Sections}

\usepackage{xspace}

\DeclareFontFamily{U}{MnSymbolC}{}
\DeclareSymbolFont{MnSyC}{U}{MnSymbolC}{m}{n}
\DeclareFontShape{U}{MnSymbolC}{m}{n}{
    <-6>  MnSymbolC5
   <6-7>  MnSymbolC6
   <7-8>  MnSymbolC7
   <8-9>  MnSymbolC8
   <9-10> MnSymbolC9
  <10-12> MnSymbolC10
  <12->   MnSymbolC12%
}{}
\DeclareMathSymbol{\powerset}{\mathord}{MnSyC}{180}

\begin{document}

\title{Joint PoS Tagging and Stemming for Agglutinative Languages}

\author{Necva B{\"{o}}l{\"{u}}c{\"{u}} \and Burcu Can }

\institute{
Department of Computer Engineering, Hacettepe University\\Beytepe, Ankara, 06800, Turkey\\
\email{\{necva,burcu\}@cs.hacettepe.edu.tr}
}

\maketitle

\begin{abstract}
The number of word forms in agglutinative languages is theoretically infinite and this variety in word forms introduces sparsity in many natural language processing tasks. Part-of-speech tagging (PoS tagging) is one of these tasks that often suffers from sparsity. In this paper, we present an unsupervised Bayesian model using Hidden Markov Models (HMMs) for joint PoS tagging and stemming for agglutinative languages. We use stemming to reduce sparsity in PoS tagging. Two tasks are jointly performed to provide a mutual benefit in both tasks. Our results show that joint POS tagging and stemming improves PoS tagging scores. We present results for Turkish and Finnish as agglutinative languages and English as a morphologically poor language. 
\end{abstract}

\section{Introduction}\label{sec:intro}
Part-of-speech (PoS) tagging is one of the essential tasks in many natural language processing (NLP) applications, such as machine translation, sentiment analysis, question answering etc. The task is especially crucial for the disambiguation of a word. For example, the word \textit{saw} can correspond to either a noun or a verb. The meaning is ambiguous unless its syntactic category is known. Once its syntactic category is assigned a noun, it becomes clear that the word corresponds to the tool, \textit{saw}. 

Agglutinative languages introduce the sparsity problem in NLP tasks due to their rich morphology. Hankamer \cite{Hankamer} claims that the number of various word forms in an agglutinative language like Turkish is theoretically infinite. The sparsity emerges with out-of-vocabulary (OOV) problem and is often a bottleneck in PoS tagging. Therefore, PoS tagging in agglutinative languages becomes even more challenging compared to other languages with a poorer morphology. 

In this paper, we tackle the sparsity problem by combining PoS tagging with stemming in the same framework by reducing the number of distinct word forms to distinct stem types. Stemming is the process of finding the stem of the word by removing its suffixes. In stemming, normally inflectional suffixes are stripped off, whereas the derivational suffixes are kept because the stem refers to a different word type (i.e. lemma). For example, the stem of \textit{bookings} is \textit{booking} since \textit{-s} is an inflectional suffix, whereas \textit{-ing} is a derivational suffix. Moreover, \textit{booking} exists in dictionary as a word itself.

Many PoS tagging models ignore the morphological structure of the agglutinative languages. In this paper, we present an unsupervised model for PoS tagging that jointly finds stems and PoS tags. We propose different approaches to the same model, where all of them learn the tags and stems from a given raw text in a fully unsupervised setting. Different approaches show that using stems rather than words in learning PoS tagging improves PoS tagging performance, which also helps in learning stems cooperatively. Our model is based on a Bayesian hidden Markov Model (HMM) with a second order Markov chain for the tag transitions. We test with different emission types and the results show that emitting stems rather than words improves PoS tagging accuracy. 

The paper is organized as follows: Section \ref{related_work} addresses the related work on unsupervised POS tagging and stemming, section \ref{model} describes our Bayesian HMM model and the different settings of the same Bayesian model applied for joint learning of PoS tags and stems, section \ref{inference} explains the inference algorithm to learn the model, section \ref{experiments} presents the experimental results obtained from different datasets for English, Turkish and Finnish languages along with a discussion on the results, and finally section \ref{conclusion} concludes the paper with the future goals.

\section{Related Work}
\label{related_work}

\subsection{PoS Tagging}
Various methods have been applied for PoS tagging. Some of them have seen PoS tagging as a clustering/classification problem. Brown et al.~\cite{brown1992class} introduce a class-based n-gram model that learns either syntactic or semantic classes of words depending on the adopted language model; Sch{\"u}tze~\cite{schutze1993part} classifies the vector representation of words using neural networks to learn syntactic categories; Clark~\cite{clark2000inducing} proposes a probabilistic context distributional clustering to cluster words occurring in similar contexts, thereby having similar syntactic features. Bienmann~\cite{biemann2006unsupervised} introduces a graph clustering algorithm as a PoS tagger. The graph based tagger involves two stages: In the first stage, words are clustered based on their contextual statistics; in the second stage, less frequent words are clustered using their similarity scores. 

Some other approaches have tackled PoS tagging as a sequential learning problem. For that purpose, hidden Markov models (HMMs) are commonly used for PoS tagging. HMM-based PoS tagging models go back to Merialdo~\cite{merialdo1994tagging}. Merialdo uses a trigram HMM model with maximum likelihood (ML) estimation. Trigrams'n'Tags (TnT)~\cite{brants2000tnt} is another statistical PoS tagger that uses a second order Markov Model with also maximum likelihood estimation. 

 Two tasks are jointly performed to provide a mutual benefit in bJohnson~\cite{johnson2007doesn} compares the estimators used in HMM PoS taggers. He discovers that Expectation-Maximization (EM) is not good at estimation in HMM-based PoS taggers. Gao et al.~\cite{gao2008comparison} also compare different Bayesian estimators for HMM PoS taggers. Gao et al. state that Gibbs sampler performs better on small datasets with few tags, but variational Bayes performs better on larger datasets.

Bayesian methods have also been used in PoS tagging. Goldwater and Griffiths~\cite{goldwater2007fully} adopt Bayesian learning in HMMs. HMM parameters are modeled as a Multinomial-Dirichlet distribution. In this paper, we also use their model as a baseline to our joint model. 


Van Gael et al.~\cite{van2009infinite} and Synder et al.~\cite{snyder2008unsupervised} introduce infinite HMMs that are non-parametric Bayesian where the number of states is not set and the states can grow with data. 


\subsection{Stemming}

Stemmers are mainly based on three approaches: rule-based, statistical, and hybrid.


Rule based stemmers, as the name implies, extract the base forms by using manually defined rules. The oldest stemmers are rule-based \cite{lovins1968development,porter1980algorithm,porter2001snowball}. 


One of the earliest statistical stemmers is developed by Xu and Croft~\cite{xu1998corpus}. Their method makes use of co-occurences of words to deal with words grouped in equivalence classes that are built by aggressive stemming. Mayfield and McNamee~\cite{mayfield2003single} propose a language independent n-gram stemmer. In their approach, stems are induced using n-gram letter statistics obtained from a corpus. Melucci et al.~\cite{melucci2003novel} implement a HMM-based stemmer using ML estimate to select the most likely stem and suffix based on the substring frequencies obtained from a corpus.

Linguistica~\cite{goldsmith2001unsupervised}, although being an unsupervised morphological segmentation system, is also used as a stemmer. The method is based on Minimum Description Length (MDL) model that aims to minimize the size of the lexicon by segmenting words into its segments. 


GRAph-based Stemmer (GRAS) is introduced by Paik et al.~\cite{paik2011gras} that groups words to find suffix pairs. These suffix pairs are used to build an undirected graph. Another unsupervised stemmer of the same authors~\cite{paik2011novel} use co-occurence statistics of words to find the common prefixes. 

HPS~\cite{brychcin2015hps} is one of the recent unsupervised stemmers that exploits lexical and semantic information to prepare large-scale training data in the first state, and use a maximum entropy classifier in the second stage by using the training data obtained from the first stage.

Hybrid stemmers involve different methods in a single model. Popat et al. \cite{popat2010hybrid} propose a hybrid stemmer for Gujarati that combines  statistical and rule-based models. MAULIK~\cite{mishra2012maulik} introduce another  hybrid stemmer that combines the rule-based stemmers. A word is searched in the lexicon. If not found, suffix stripping rules are used to detect the stem. Adam et al. \cite{adam2010efficient} apply PoS tagging and then use a rule-based stemmer to strip off the suffix from the word based on its tag in a pipelineframework. 


\newpage

\section{Model \& Algorithm}
\label{model}

We define a joint PoS tagger and stemmer that extends the fully Bayesian PoS tagger by Goldwater and Griffiths~\cite{goldwater2007fully}. By joining PoS tagging and stemming, we aim to reduce the sparsity in PoS tagging for agglutinative languages while also improving the stemming accuracy using the tag information. 

\subsection{Word-based Bayesian HMM Model}

\begin{figure*}[t]
\label{word-based}
\centering
\includegraphics[clip, trim=0cm 6cm 0.5cm 3cm, scale=0.25]{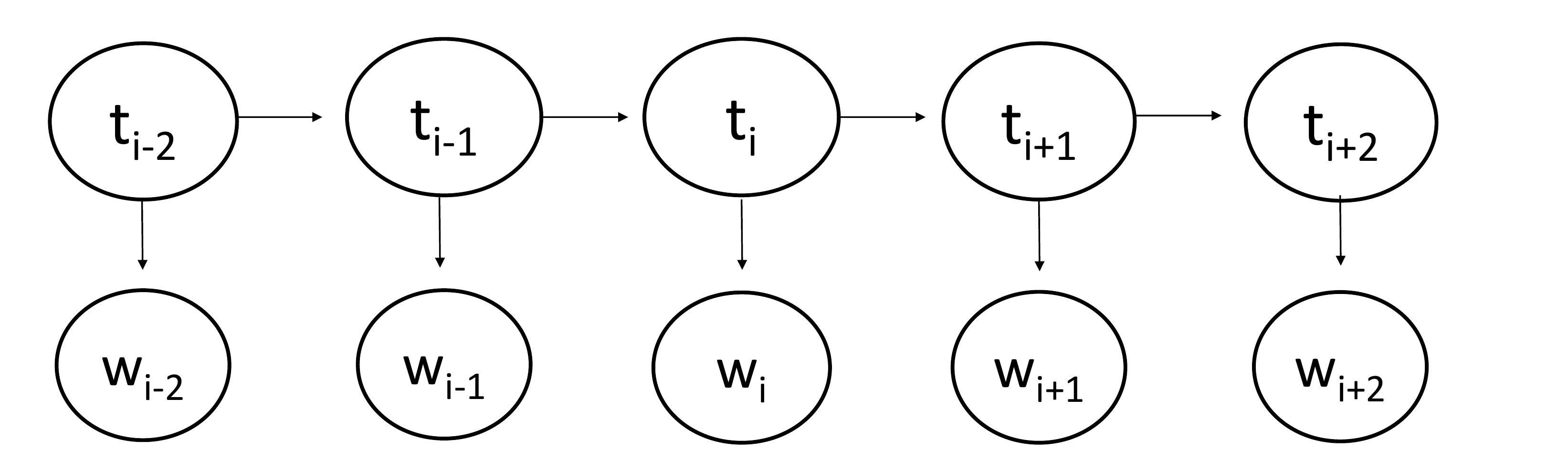}
\caption{Word-based HMM Model for PoS tagging}
\end{figure*}

The word-based Bayesian HMM model (Goldwater and Griffiths \cite{goldwater2007fully}) for PoS tagging is defined as follows (see Fig.~\ref{word-based}):
\begin{eqnarray}\label{eq:eq1}
t_{i} | t_{i-1}=t , \tau^{(t,t^{'})} &\propto& Mult(\tau^{(t,t^{'})}) \\
w_{i} | t_{i}=t , \omega^{(t)} &\propto& Mult(\omega^{(t)}) \\
\tau^{(t,t^{'})} | \alpha &\propto& Dirichlet(\alpha) \\
\omega^{(t)} | \beta &\propto& Dirichlet(\beta) 
\end{eqnarray}
where $w_{i}$ denotes the ith word in the corpus and $t_{i}$ is its tag. $Mult(\omega^{t})$ is the emission distribution in the form of a Multinomial distribution with parameters $\omega^{(t)}$ that are generated by $Dirichlet(\beta)$ with hyperparameters $\beta$. Analogously, $Mult(\tau^{(t,t^{'})})$ is the transition distribution with parameters $\tau^{(t,t^{'})}$ that are generated by $Dirichlet(\alpha)$ with hyperpameters $\alpha$.


The conditional distribution of $t_i$ under this model is:
\begin{eqnarray}\label{eq:eq2}
P(t_{i}|t_{-i},w,\alpha,\beta) &=&
\frac{n_{(t_{i},w_{i})}+\beta}{n_{t_{i}}+W_{t_{i}}\beta} 
 \cdot\frac{n_{(t_{i-1},t_{i})}+\alpha}{n_{t_{i-1}}+T\alpha}  \\ \nonumber
 &\cdot& \frac{n_{(t_{i},t_{i+1})+I(t_{i-1}=t_{i}=t_{i+1})}+\alpha}{n_{t_{i}}+I(t_{i-1}=t_{i})+T\alpha} 
\end{eqnarray}
where $W_{t_{i}}$ is the number of word types in the corpus, $T$ is the size of the tag set, $n_{t_{i}}$ is the number  of words tagged with $t_i$, $n_{(t_{i-1},t_{i})}$ is the frequency of tag bigram $<t_{i-1},t_{i}>$. $I(.)$ is a function that gives 1 if its argument is true, and otherwise 0.

\subsection{Stem-based Bayesian HMM Model}

We extend the basic HMM model for PoS tagging introduced by Goldwater and Griffiths~\cite{goldwater2007fully} by replacing the word emissions with stem emissions in order to reduce the emission sparsity, thereby mitigating the size of the out-of-vocabulary words. Therefore, we obtain a joint PoS tagger and stemmer with this model.

\begin{figure*}[t]
\label{stem-based}
\centering
\includegraphics[clip, trim=0cm 6cm 0.5cm 3cm, scale=0.25]{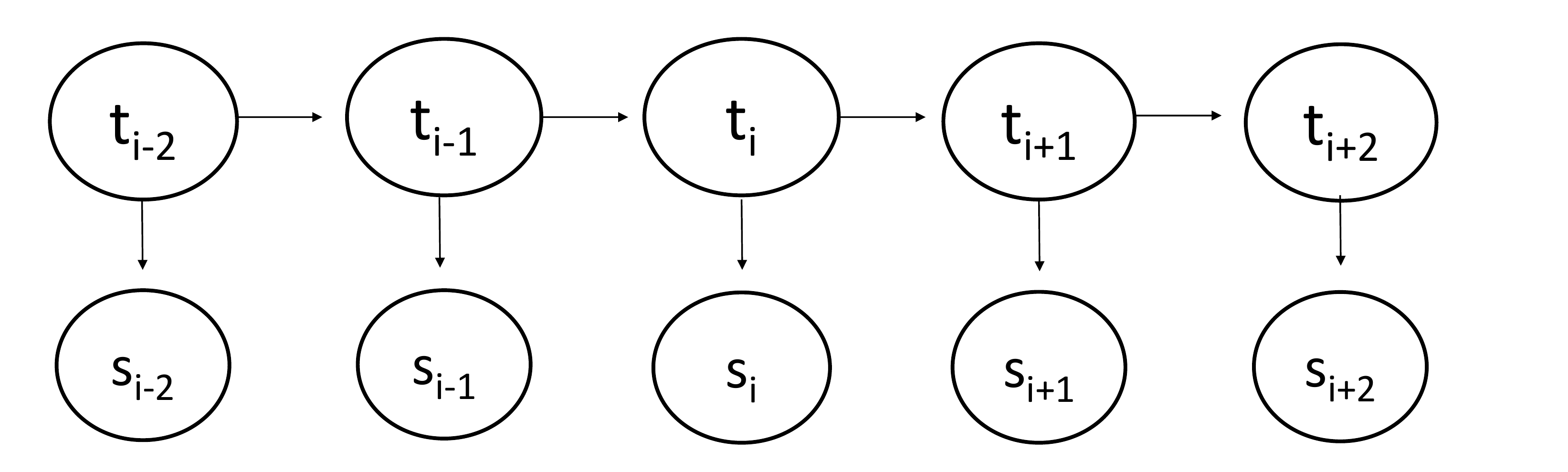}
\caption{Stem-based HMM Model for PoS Tagging}
\end{figure*}

The stem-based model is defined as follows (see Fig.~\ref{stem-based}):
\begin{eqnarray}\label{eq:eq3}
t_{i} | t_{i-1}=t , \tau^{(t,t^{'})} &\propto& Mult(\tau^{(t,t^{'})}) \\
s_{i} | t_{i}=t , \omega^{(t)} &\propto& Mult(\omega^{(t)}) \\
\tau^{(t,t^{'})} | \alpha &\propto& Dirichlet(\alpha) \\
\omega^{(t)} | \beta &\propto& Dirichlet(\beta) 
\end{eqnarray}
Here, $t_{i}$ and $s_{i}$ are the \textit{i}th tag and stem, where $w_i=s_i+m_i$, $m_i$ being the suffix of $w_i$.

Under this model, the conditional distribution of $t_i$ becomes as follows:
\begin{eqnarray}\label{eq:eq4}
P(t_{i}|t_{-i},s_i,\alpha,\beta) &=&
\frac{n_{(t_{i},s_{i})}+\beta}{n_{t_{i}}+S_{t}\beta}
 \cdot \frac{n_{(t_{i-1},t_{i})}+\alpha}{n_{t_{i-1}}+T\alpha} \\ \nonumber
 &\cdot& \frac{n_{(t_{i},t_{i+1})+I(t_{i-1}=t_{i}=t_{i+1})}+\alpha}{n_{t_{i}}+I(t_{i-1}=t_{i})+T\alpha} 
\end{eqnarray}
where $S_{t}$ is the number of stem types in the corpus. When compared to the word-based model, the number of word types reduces to stem types. Therefore, sparsity also decreases.

\subsection{Stem/Suffix-based Bayesian HMM Model}
\begin{figure*}[t]
\label{stem-suffix}
\centering
\includegraphics[clip, trim=0cm 6.5cm 0.3cm 3cm, scale=0.27]{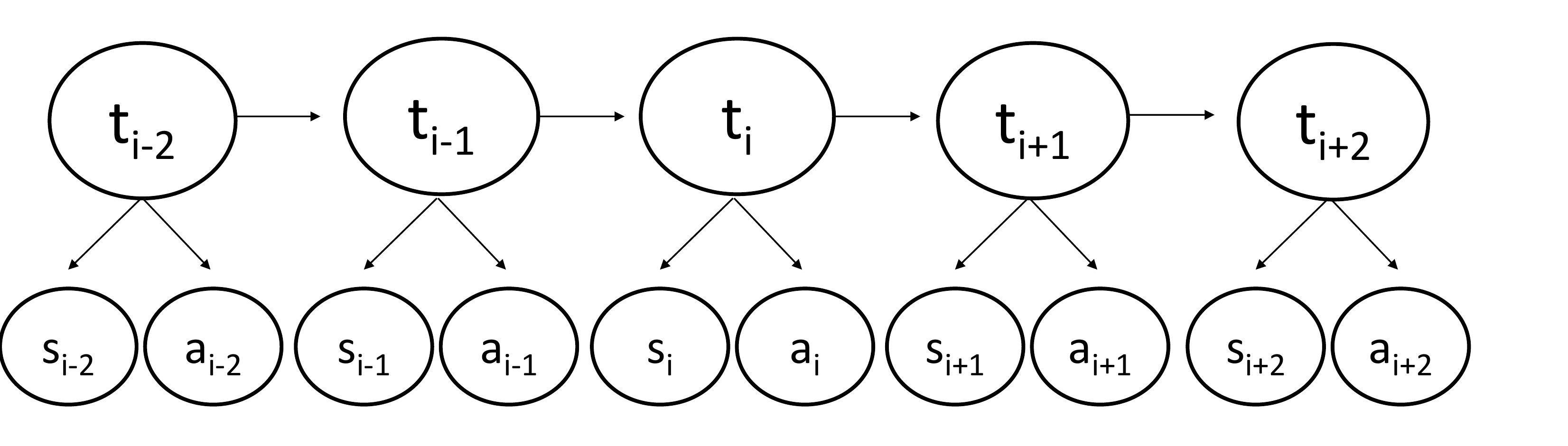}
\caption{Stem/suffix-based HMM Model for PoS tagging}
\end{figure*}

Words belonging to the same syntactic category take also similar suffixes. For example, words ending with \textit{ly} are usually adverbs, whereas words ending with \textit{ness} are usually nouns. We include suffixes in the emissions in addition to the stems (see Fig.~\ref{stem-suffix}): 
\begin{eqnarray}\label{eq:eq5}
t_{i} | t_{i-1}=t , \tau^{(t,t^{'})} &\propto& Mult(\tau^{(t,t^{'})}) \\
s_{i} | t_{i}=t , \omega^{(t)} &\propto& Mult(\omega^{(t)}) \\
m_{i} | t_{i}=t , \psi^{(t)} &\propto& Mult(\psi^{(t)}) \\
\tau^{(t,t^{'})} | \alpha &\propto& Dirichlet(\alpha) \\
\omega^{t} | \beta &\propto& Dirichlet(\beta) 			\\
\psi^{(t)} | \gamma &\propto& Dirichlet(\gamma) 
\end{eqnarray}
where $m_{i}$ is the suffix of $w_i=s_i+m_i$ which is generated by $Mult(\psi^{(t)})$ with parameters drawn from $Dirichlet(\gamma)$ with hyperparameters $\gamma$.

The new conditional distribution of $t_i$ becomes:
\begin{eqnarray}\label{eq:eq6}
P(t_{i}|t_{-i},s_i,m_i,\alpha,\beta,\gamma) =
& & \frac{n_{(t_{i},s_{i})}+\beta}{n_{t_{i}}+S_{t}\beta} \cdot \frac{n_{(t_{i},m_{i})}+\gamma}{n_{t_{i}}+M_{t}\gamma}
\cdot\frac{n_{(t_{i-1},t_{i})}+\alpha}{n_{t_{i-1}}+T\alpha} \nonumber \\ 
& & \cdot\frac{n_{(t_{i},t_{i+1})+I(t_{i-1}=t_{i}=t_{i+1})}+\alpha} {n_{t_{i}}+I(t_{i-1}=t_{i})+T\alpha} 
\end{eqnarray}
where $M_{t}$ is the number of suffix types in the corpus. 

\section{Inference}
\label{inference}
We use Gibbs sampling~\cite{geman1984stochastic} for the inference of the model. $\textbf{t}$ are drawn from the posterior distribution $P(\textbf{t}|\textbf{w},\alpha,\beta) \propto P(\textbf{w}|\textbf{t},\beta)P(\textbf{t}|\alpha)$ in the word-based Bayesian HMM model, $P(\textbf{t}|\textbf{s},\alpha,\beta) \propto P(\textbf{s}|\textbf{t},\beta)P(\textbf{t}|\alpha)$ in the stem-based model, and $P(\textbf{t}|\textbf{s},\textbf{m},\alpha,\beta,\gamma) \propto P(\textbf{s}|\textbf{t},\beta)P(\textbf{m}|\textbf{t},\gamma)P(\textbf{t}|\alpha)$ in the stem/suffix-based model. 
 
In the word-based Bayesian HMM model, all tags are randomly initialized at the beginning of the inference. Then each word's tag is sampled from the model's posterior distribution given in Equation~\ref{eq:eq2}. This process is repeated until the system converges. 

In the stem-based and stem/suffix-based model, all tags are randomly initialized and all words are split into two segments randomly. In each iteration of the algorithm, a tag and a stem are sampled for each word from the posterior distribution given in Equation~\ref{eq:eq4} and Equation~\ref{eq:eq6} respectively.

\newpage

\section{Experiments \& Results}
\label{experiments}
\subsubsection{Data:}
We used three datasets for the experiments and evaluation:
\begin{itemize}
\item \textit{Turkish:} METU-Sabanc{\i} Turkish Treebank~\cite{oflazer2003building} that consists of 53751 word tokens.
\item \textit{English:} The first 12K and 24K words from the WSJ Penn Treebank~\cite{marcus1993building}.
\item \textit{Finnish:} The first 12k and 24k words from FinnTreeBank corpus that is a revised version of the original FTB1\footnotemark[1].
\end{itemize}

\footnotetext[1]{Available at \url{http://www.ling.helsinki.fi/kieliteknologia/tutkimus/treebank/sources/}}

There are 41 tags in both Penn Treebank and METU-Sabanc{\i} Turkish Treebank, and 12 tags in FinnTreeBank. We mapped the three tagsets to the Universal tagset~\cite{petrov2011universal} that involves 12 categories. The new tagsets for Turkish, English and Finnish are given in Table~\ref{tab:reducedPenn} and Table~\ref{tab:reducedMetu}. Therefore, the size of the tagset is 12 in all experiments for three languages. 

\begin{table}[t!]
\caption{The mapping of the Universal tagset to the Penn Treebank tagset and the FinnTreeBank tagset}
\centering
\label{my-label}
\begin{tabular}{lll}
\hline
Universal tagset & Penn Treebank tagset   &   FinnTreeBank tagset                  \\ \hline
VERB             & VBP,VBD,VBG,VBN,VB,VBZ,MD   & V                \\
PRON             & WP,PRP$,PRP,WP$              & Pron                 \\
PUNCT            & (``),(,),-LRB-,-NONE-,-RRB-,(.),(:),(''),\$ & Punct \\
PRT              & RP,TO      & Pcle                                   \\
DET              & WDT,EX,PDT,DT       &Det                        \\
NOUN             & NN,NNP,NNPS,NNS  & N                           \\
ADV              & RB,RBR,WRB,RBS    & Adv                          \\
ADJ              & JJ,JJS        & A                              \\
UNKNOWN          & FW,UH    & Symb, Foreign, Interj                                    \\
ADP              & IN           & Adp                                \\
NUM              & CD     & Num                                 \\
CONJ             & CC      & C       \\       \hline                 
\end{tabular}
\label{tab:reducedPenn}
\end{table}

\begin{table}[t!]
\caption{The mapping of the Universal tagset to the Metu-Sabanc{\i} Turkish Treebank tagset}
\centering
\label{my-label}
\begin{tabular}{ll}
\hline
Universal tagset & Metu-Sabanc{\i} Turkish Treebank tagset                                                                                                                                                                       \\ \hline
Noun             & \begin{tabular}[c]{@{}l@{}}Noun\_Pron,Noun\_Ins,Noun\_Nom,Noun\_Verb,Noun\_Loc,\\Noun\_Acc,Noun\_Abl,Noun\_Gen, Noun\_Dat,Noun\_Adj,\\Noun\_Num,Noun\_Pnon,Noun\_Postp,Noun\_Equ\end{tabular} \\
Adj              & Adj\_Noun,Adj\_Verb,Adj,Adj\_Pron,Adj\_Postp,Adj\_Num                                                                                                                                       \\
Adv              & Adv\_Verb,Adv\_Adj,Adv\_Noun,Adv                                                                                                                                                            \\
Conj             & Conj                                                                                                                                                                                        \\
Det              & Det                                                                                                                                                                                         \\
Interj           & Interj                                                                                                                                                                                      \\
Ques             & Ques                                                                                                                                                                                        \\
Verb             & Verb,Negp,Verb\_Noun,Verb\_Postp,Verb\_Adj,Verb\_Adv,Verb\_Verb                                                                                                                             \\
Postp            & Postp                                                                                                                                                                                       \\
Num              & Num                                                                                                                                                                                         \\
Pron             & Pron,Pron\_Noun                                                                                                                                                                             \\
Punc             & Punc           \\ \hline                                                                                                                                                                            
\end{tabular}
\label{tab:reducedMetu}
\end{table}

We ran each model with four settings of parameters. In the first setting, we assigned $\alpha$ =0.001, $\beta$=0.1, and $\gamma$=0.001 (indicated as setting $1$ in the tables); in the second setting, $\alpha$ =0.003, $\beta$=1, and $\gamma$=0.003 (indicated as setting $2$ in the tables); in the third setting, $\alpha$ =0.001, $\beta$=0.1, and $\gamma$=0.001 (indicated as setting $3$ in the tables); and in the fourth setting we assigned $\alpha$ =0.003, $\beta$=1, and $\gamma$=0.003 (indicated as setting $4$ in the tables). 

The stemming results are evaluated based on the accuracy measure. We compare our stemming results obtained from the stem-based Bayesian HMM (Bayesian S-HMM) and stem/suffix-based HMM (Bayesian SM-HMM) with HPS~\cite{brychcin2015hps} and FlatCat~\cite{gronroos2014morfessor}. The results for Turkish and Finnish obtained from the Metu-Sabanc{\i} Turkish Treebank and FinnTreeBank respectively are given in \cref{tab:stemResult}. Although our stemming results are far behind the results of the HPS algorithm for Turkish, our Finnish results are on a par with HPS and Morfessor FlatCat. The results show that using suffixes does not help in stemming. Using stem emissions alone gives the best accuracy for stemming in the joint task. 

Since the English stems are not covered in Penn TreeBank, we were not able to evaluate the English stemming results.

\begin{table}[t!]
\caption{Stemming results for Turkish and Finnish based on four parameter settings}
\centering
\begin{tabular}{lll l l}
\hline
& \multicolumn{4}{c}{\textbf{Accuracy (\%)}} \\ \hline
& Model & Metu& Finn 12K& Finn 24K \\ \hline
& HPS \footnotemark[3] \cite{brychcin2015hps}& \textbf{53.79}  &\textbf{28.19}&27.04\\
& Morfessor FlatCat\footnotemark[2] \cite{gronroos2014morfessor}& 52.06 &24.47&25.93 \\ \hline
\multirow{2}{*}{1}&Bayesian S-HMM & 46.21 &23.24& 22.32\\
& Bayesian SM-HMM  &34.97 &26.69& 26.40\\ \hline
\multirow{2}{*}{2}&Bayesian S-HMM &46.39  &23.49& 22.14\\
& Bayesian SM-HMM  &34.82 &27.95&\textbf{27.38} \\ \hline
\multirow{2}{*}{3}&Bayesian S-HMM & 46.57 &23.28&22.03 \\
& Bayesian SM-HMM  & 34.97&27.45&26.74 \\ \hline
\multirow{2}{*}{4}&Bayesian S-HMM & 46.46 &18.62&22.57\\
& Bayesian SM-HMM  &32.13 &24.78&24.40 \\ \hline
\end{tabular}
\label{tab:stemResult}
\end{table}

Examples to correct and incorrect stems in all languages are given in \Cref{tab:correctStem}. The results show that our joint model can find the common endings, such as \textit{s, ed, ted, er, d, e, ing} in English. However, since we do not exploit any semantic features in the model, words such as \textit{filter} can be stemmed as \textit{filt+er}. This is also one of the main problems in morphological segmentation models that rely only on the orthographic features. Our stemming results are promising, but it shows that it is not sufficient to reduce the sparsity based on the common segments and it requires more features.

\begin{table}[]
\caption{Examples to correct and incorrect stems}
\centering
\label{my-label}
\begin{tabular}{ll ll ll}
\hline
\multicolumn{2}{c}{\textbf{Turkish}}  & \multicolumn{2}{c}{\textbf{English}}  & \multicolumn{2}{c}{\textbf{Finnish}}  \\ \hline
Correct& Incorrect & Correct & Incorrect& Correct& Incorrect \\ \hline
siz-lere & {\"{o}}\u{g}re-ncilere  & year+s  & chairma+n  &  niska+an & sai+si  \\
 dur-du  & jandar+mal{\i}\u{g}{\i}na  & york-bas+ed &  repor+ted & suomenmaa+\#  &  t{\"a}nn+e   \\
g{\"{o}}r-d{\"{u}}\u{g}{\"{u}}n{\"{u}}z & rastla+d{\i}\u{g}{\i}  &  talk+ing    &  filt+er  &  pappila+ssa &    tul+ee  \\
g{\"{o}}z-leri         & iznin+e &   the+\# &    sai+d  &   piste+tt{\"a}    &    kotii+n  \\
abone+\#         & geti+riliyor                  & inform+ation   &  institut+e                   &  valinta+nsa      &    oll+a     \\ \hline         
\end{tabular}
\label{tab:correctStem}
\end{table}

We evaluate PoS tagging results with many-to-one accuracy and variation of information (VI) measure~\cite{rosenberg2007v}. Turkish, English and Finnish results are given in \Cref{tab:evaluateTurkish},  \Cref{tab:evaluateEnglish}, and \Cref{tab:evaluateFinnish} respectively. The overall results show that using stems rather than words leads to better results in three languages. Therefore, the Bayesian S-HMM model outperforms other two models in three languages in general. Although English has got a poor morphology when compared to Turkish and Finnish, the Bayesian S-HMM model still outperforms other two models. Using suffixes also does not help in PoS tagging and its scores are generally behind the Bayesian S-HMM model. However, in some parameter settings Bayesian SM-HMM model outperforms other two Bayesian models.

The overall PoS tagging results show that our stem-based and stem/suffix-based Bayesian models outperform both Brown Clustering~\cite{brown1992class} and word-based Bayesian HMM model \cite{goldwater2007fully} for three languages according to both many-to-one measure and VI measure. 

\begin{table}[t!]
\caption{POS tagging evaluation results for Turkish}
\centering
\begin{tabular}{ll ll}
& Model   & \multicolumn{2}{l}{\textbf{Metu} }  \\ \hline
&  & \textbf{Many-to-1}           & \textbf{VI}\\ \hline
\multirow{3}{*}{1} & Bayesian HMM   & 57.58            & 10.58         \\
& Bayesian S-HMM   &      55.34       & 10.57     \\
 & Bayesian SM-HMM    &     56.17        &    \textbf{10.47}      \\ \hline
 \multirow{3}{*}{2} & Bayesian HMM   & 56.56            & 11.01        \\
& Bayesian S-HMM   &     \textbf{57.70}      & 10.48     \\
 & Bayesian SM-HMM    &    55.30         &  10.60        \\ \hline
 \multirow{3}{*}{3} & Bayesian HMM   & 55.89            & 10.70         \\
& Bayesian S-HMM   &      54.99       &  10.59    \\
 & Bayesian SM-HMM    &       57.08      &   \textbf{10.47}       \\ \hline
 \multirow{3}{*}{4} & Bayesian HMM   & 56.64            & 10.94         \\
& Bayesian S-HMM   &        57.01     &  10.48    \\
 & Bayesian SM-HMM    &       55.99      &   10.58       \\ \hline

& Brown Clustering \footnotemark[4] \cite{brown1992class} & 54.91            & 10.83     \\ \hline
\end{tabular}
\label{tab:evaluateTurkish}
\label{tab:evaluateTurkish}
\label{my-label}
\end{table}


\begin{table}[t!]
\caption{PoS tagging evaluation results for English}
\centering
\label{my-label}
\begin{tabular}{ll ll | ll}
                   & Model                                                         & \multicolumn{2}{l|}{\textbf{Penn 12k}} & \multicolumn{2}{l}{\textbf{Penn 24k }} \\ \hline
                   &                                                               & \textbf{Many-to-1}         & \textbf{VI}        & \textbf{Many-to-1}           & \textbf{VI}            \\ \hline
\multirow{3}{*}{1} & Bayesian HMM & 49.10    & 7.80  &52.34    & 7.94    \\
 & Bayesian S-HMM    &50.84  &\textbf{7.54}         & 49.76   & 7.92   \\
& Bayesian SM-HMM &  51.04             &  7.86      & 51.65 & 8.12   \\ \hline
\multirow{3}{*}{2} & Bayesian HMM & 41.88    & 8.46  &47.13    & 8.52    \\
 & Bayesian S-HMM    & \textbf{54.92}  & 7.58          & 54.52   & 7.86   \\
& Bayesian SM-HMM &  48.76             & 7.92       & 51.93 &  8.13  \\ \hline
\multirow{3}{*}{3} & Bayesian HMM & 50.70    & 7.75 &46.74    & 8.15    \\
 & Bayesian S-HMM    & 50.57  & 7.71          & 52.05   & 7.97   \\
& Bayesian SM-HMM &   53.06            & 7.64       & 51.39 &  7.95  \\ \hline
\multirow{3}{*}{4} & Bayesian HMM & 45.15    & 8.33  &45.65    & 8.55    \\
 & Bayesian S-HMM    & 52.67   & 7.69          & \textbf{55.32}   & \textbf{7.67}   \\
& Bayesian SM-HMM &       53.15        &  7.67      & 51.41 & 8.17    \\ \hline
& Brown Clustering \footnotemark[4] \cite{brown1992class} & 53.78             & 7.58          & 54.11             & 7.78      \\ \hline
\end{tabular}
\label{tab:evaluateEnglish}
\end{table}


\begin{table}[t!]
\caption{PoS tagging evaluation results for Finnish}
\centering
\label{my-label}
\begin{tabular}{ll ll | ll}
& Model & \multicolumn{2}{l|}{\textbf{FinnTreeBank 12k }} & \multicolumn{2}{l}{\textbf{FinnTreeBank 24k }} \\ \hline
 &  & \textbf{Many-to-1}  & \textbf{VI}  & \textbf{Many-to-1} & \textbf{VI}  \\ \hline
\multirow{3}{*}{1} & Bayesian HMM & 42.43   & 10.65  &  44.96    & 11.43         \\
 & Bayesian S-HMM    &47.29 &10.39         & 46.83   &   11.16   \\
& Bayesian SM-HMM &   48.84         & 10.27      &  48.45          &   11.09     \\ \hline
\multirow{3}{*}{2} & Bayesian HMM & 42.94   & 10.74  &  44.50    & 11.56         \\
 & Bayesian S-HMM    &\textbf{51.15} &10.34         & 51.27   &    11.10  \\
& Bayesian SM-HMM &    48.66        & 10.32       &  49.04          &  11.12      \\ \hline
\multirow{3}{*}{3} & Bayesian HMM & 42.57   & 10.61 &  46.47    & 11.28         \\
 & Bayesian S-HMM    &45.91 &10.36         & 51.58   &    11.07  \\
& Bayesian SM-HMM &     49.40       & \textbf{10.25}     &48.43            &    11.08    \\ \hline
\multirow{3}{*}{4} & Bayesian HMM & 42.96   & 10.75  &  44.27    & 11.58         \\
 & Bayesian S-HMM    &50.47 &10.38         &  \textbf{51.52}  & \textbf{11.02}    \\
& Bayesian SM-HMM &      49.02      &10.31       &  48.43          &   11.14     \\ \hline
& Brown Clustering \footnotemark[4] \cite{brown1992class} & 44.33              & 10.64         & 47.95            & 11.33   \\ \hline
\end{tabular}
\label{tab:evaluateFinnish}
\end{table}
\footnotetext[2]{Morfessor FlatCat: \url{https://github.com/aalto-speech/flatcat}}
\footnotetext[3]{HPS: \url{http://liks.fav.zcu.cz/HPS/}}
\footnotetext[4]{Brown Clustering: \url{http://www.cs.berkeley.edu/~pliang/software/brown-cluster-1.2.zip}(Percy Liang)}

\section{Conclusion \& Future Work}
\label{conclusion}
In this paper, we extend the Bayesian HMM model~\cite{goldwater2007fully} for joint learning of PoS tags and stems in a fully unsupervised framework. Our model reduces the sparsity by using stems and suffixes instead of words in a HMM model. The results show that using stems and suffixes rather than words outperforms a simple word-based Bayesian HMM model for especially agglutinative languages such as Turkish and Finnish. Although English has got a poor morphology, the English PoS tagging results are also better when the stems are used instead of words. 

Although our Turkish stemming results are far behind the other compared models, our Finnish stemming results are on par with other models.

We aim to use other features (such as semantic features) in our model to capture the semantic similarity between the stems and their derived forms, which is left as a future work. 

Our model does not deal with irregular word forms. We also leave this as a future work. 


\section*{Acknowledgments}
This research is supported by the Scientific and Technological Research Council of Turkey (TUBITAK) with the project number EEEAG-115E464.

\bibliographystyle{splncs03}
\bibliography{paper}


\end{document}